\documentclass[10pt,twocolumn,letterpaper]{article}

\usepackage{iccv}
\usepackage{times}
\usepackage{epsfig}
\usepackage{graphicx}
\usepackage{amsmath}
\usepackage{amssymb}

\usepackage{algorithm}
\usepackage[noend]{algpseudocode}
\algrenewcommand\algorithmicdo{}
\algnewcommand\algorithmicto{\textbf{to}}
\algdef{SE}[FOR]{ForTo}{EndFor}[2]{\algorithmicfor\ #1\ \algorithmicto\ #2\ \algorithmicdo}{\algorithmicend\ \algorithmicfor}
\usepackage[subrefformat=parens]{subcaption}
\DeclareMathOperator*{\argmin}{arg\,min}
\newcommand{\CC}{C\nolinebreak\hspace{-.05em}\raisebox{.4ex}{\scriptsize\bf +}\nolinebreak\hspace{-.10em}\raisebox{.4ex}{\scriptsize\bf +}}
\def\CC{{C\nolinebreak[4]\hspace{-.05em}\raisebox{.4ex}{\scriptsize\bf ++}}}
\usepackage[font=small]{caption}
\usepackage[font=small]{subcaption}
\usepackage{url}

\usepackage[breaklinks=true,bookmarks=false]{hyperref}

\iccvfinalcopy 

\def\iccvPaperID{****} 

\ificcvfinal\fi

\begin{document}

\title{Fast and Flexible Image Blind Denoising via Competition of Experts}

\author{Shunta Maeda\\
Navier Inc.\\
{\tt\small shunta@navier.co.jp}
}

\maketitle

\begin{abstract}
Fast and flexible processing are two essential requirements for a number of practical applications of image denoising.
Current state-of-the-art methods, however, still require either high computational cost or limited scopes of the target.
We introduce an efficient ensemble network trained via a competition of expert networks, as an application for image blind denoising.
We realize automatic division of unlabeled noisy datasets into clusters respectively optimized to enhance denoising performance.
The architecture is scalable, can be extended to deal with diverse noise sources/levels without increasing the computation time.
Taking advantage of this method, we save up to approximately $90\%$ of computational cost without sacrifice of the denoising performance compared to single network models with identical architectures.
We also compare the proposed method with several existing algorithms and observe significant outperformance over prior arts in terms of computational efficiency.
\end{abstract}

\section{Introduction}
While a lot of image denoising algorithms have been developed over the past decades, blind removal of real noises still remains challenging.
Blind denoising aims to recover a clean image from a noisy one without referring to any {\it a priori} information about the noise.
This restriction can be widely seen as a realistic requirement of practical use cases.
Fast processing is also important because the denoising often constitutes a crucial part of pre-processing pipelines in many vision tasks~\cite{tang2010deep, diamond2017dirty}.
However, current state-of-the-art methods still require either high computational cost or limited scopes of the target, unfitting for many real-world applications.

Image denoising methods are broadly classified into two categories: image prior based and discriminative learning based methods.
The image prior based methods, which explicitly define models of noises, have achieved remarkable results~\cite{buades2005non, dabov2007image, beck2009fast, gu2014weighted}.
Nevertheless, their recovering performances are limited because noise models are usually introduced manually.
In contrast, discriminative learning based methods learn the underlying mapping between clean images and noisy ones by exploiting large modeling capacity of deep convolutional neural networks (CNNs)~\cite{zhang2017beyond, zhang2018ffdnet, guo2018toward}.
Although learning based methods demonstrate advanced performances if the evaluation data resembles the training data used, they are often outperformed by image prior based methods when tested on actual noisy images~\cite{plotz2017benchmarking, SIDD_2018_CVPR}.
This drawback of learning based methods is mainly due to a lack of high-quality denoising dataset.
Some notable recent works have successfully produced such high-quality denoising dataset, for example, by systematic procedures estimating ground-truth images of noisy ones~\cite{SIDD_2018_CVPR} or to create real noisy images utilizing generative adversarial networks~\cite{Chen_2018_CVPR}.

Even if ideal pairs of clean images and noisy ones are available, learning based methods are usually inferior to image prior based methods in terms of adaptability for data outside the training domain.
A possible solution overcoming this limitation is preparing a vast amount of clean-noisy image pairs across a variety of training domains.
However, a large scale network is required to handle such dataset, resulting in a high computational cost.
Taking into account the above, one natural approach would be this: training a specialized expert for each representative domain in given dataset and selecting a few suitable experts for the evaluation.
Such an approach is widely known as mixture-of-experts (MoE)~\cite{jacobs1991adaptive, jordan1994hierarchical}.
An underlying difficulty of this approach is that real noisy images are usually unlabeled. 
Moreover, even though they are labeled, MoE trained with given labels is not guaranteed to minimize the task-oriented loss.
Thus an unsupervised clustering method minimizing the loss of the denoising with MoE is desired.

In this work, we propose a stable method to train specialized experts via competition.
The training procedure is designed for low-level vision tasks, and the network architecture is composed for fast processing applications.
We apply the method to image blind denoising and realize automatic division of unlabeled noisy datasets into appropriate domains.
To evaluate the effectiveness of the proposed method, we compare it with single network models using identical architecture.
Our extensive experiments reveal how the performances vary according to the number of experts and the size of the expert network.
In addition, we examine perceptual quality along with various noise levels.
We also compare the proposed method with several existing algorithms and observe significant outperformance in terms of computational efficiency.

We highlight the notable contributions of this work:
\begin{itemize}
  \item By utilizing our competition training method, unlabeled noisy dataset is automatically divided into appropriate clusters which are optimized respectively to enhance denoising performance.
  \item The proposed method is up to 10$\times$ faster than single network baseline without sacrifice of the denoising performance (Fig.~\ref{fig:efficiency}).
  \item The proposed method is 5$\times$ faster than BM3D, which is a state-of-the-art patch-based algorithm, when comparing at similar denoising performance (Tab.~\ref{tab:compare_awgn}).
  \item Our formulation is applicable to a variety of other low-level vision tasks with different model networks and loss functions.
\end{itemize}

\section{Related Work}
\subsection{Image Prior Based Denoising}
Image priors are utilized in most of traditional denoising methods.
Among them, patch-based denoising methods, such as non-local means (NL-means)~\cite{buades2005non}, BM3D~\cite{dabov2007image}, and WNNM~\cite{gu2014weighted} have achieved remarkable results.
These methods explicitly define noise models based on {\it a priori} knowledge, and thus they are applicable to denoising problems with unknown noise levels.
BM3D is often referred to as a benchmark because it is still one of the best denoising algorithms in terms of quality and computation time~\cite{SIDD_2018_CVPR}.
At first in BM3D algorithm, image fragments similar to a target patch are grouped into three-dimensional arrays.
After that, the arrays are processed by collaborative filtering and then transformed back into the original two-dimensional form.
While BM3D is one of the fastest patch-based algorithms, patch-based algorithms are generally time-consuming.
In addition, its recovering performance is limited due to heuristically defined noise models.

\subsection{Discriminative Learning Based Denoising}
Learning based methods, which exploit large modeling capacity of deep CNNs, have significantly improved denoising performances by learning the underlying mapping of clean-noisy image pairs.
In particular, DnCNN~\cite{zhang2017beyond} achieves state-of-the-art results by applying residual learning for a deep CNN with batch normalization.
DnCNN is often referred to as a benchmark of learning based methods due to its simple structure and impressive performance.
However, their performances are not robust in many cases to the data outside training domains and are often outperformed by image prior based methods when applied to real datasets of noisy images~\cite{SIDD_2018_CVPR}.
Several works have been conducted to relax the above limitation by feeding information about the noise level to the denoising network.
FFDNet~\cite{zhang2018ffdnet} non-blindly removes additive white Gaussian noise (AWGN) with a tunable noise level map and is one of the most efficient networks.
CBDNet~\cite{guo2018toward} can blindly remove actual noises by estimating the noise levels through a noise estimation subnetwork.
Nevertheless, flexibility and computational efficiency of these networks are still inadequate because they feature a single denoising network.

\subsection{Mixture of Experts}
MoE is an approach that replaces a single network by a weighted sum of expert networks for further improvement of prediction performance~\cite{jacobs1991adaptive, jordan1994hierarchical}. In MoE, a trainable gating network plays a pivotal role; it activates experts and computes which experts to be weighted and how much.
Instead of activating all experts uniformly, selective activation of necessary experts at the gate enables to improve the network efficiency even more~\cite{bengio2013estimating, shazeer2017outrageously}.
MoE is effective if given dataset can be classified into several different source domains.
However, appropriate partitioning of unlabeled datasets into subsets for each expert remains an open problem, and the subject relates to unsupervised domain adaptation~\cite{guo2018multi}.

\subsection{Competition of Experts}
Some recent studies have employed competition of experts \cite{lee2016stochastic, parascandolo2017learning}.
Lee \etal~\cite{lee2016stochastic} proposed Stochastic Multiple Choice Learning (sMCL) which minimizes oracle loss in ensembles of experts via competition of experts.
They applied sMCL for image classification, segmentation, and caption generation tasks.
While all experts are assigned for the prediction in sMCL, we use a single expert for a prediction to save the computational cost.
Moreover, the expert initialization and the data sampling method are also different.
In the context of causality, Parascandolo \etal~\cite{parascandolo2017learning} proposed an algorithm to recover inverse mechanisms from transformed data points through competition of experts.
Their algorithm implicitly assumes that the training data is generated by known transformations, the series of procedures referred to in their paper, whose computations of transformation are distinctly different from each other.
Thus it does not fit with a number of practical applications.

\section{Method}
\subsection{Formulation}
\begin{figure}[t]
\centering
\includegraphics[width=0.94\linewidth]{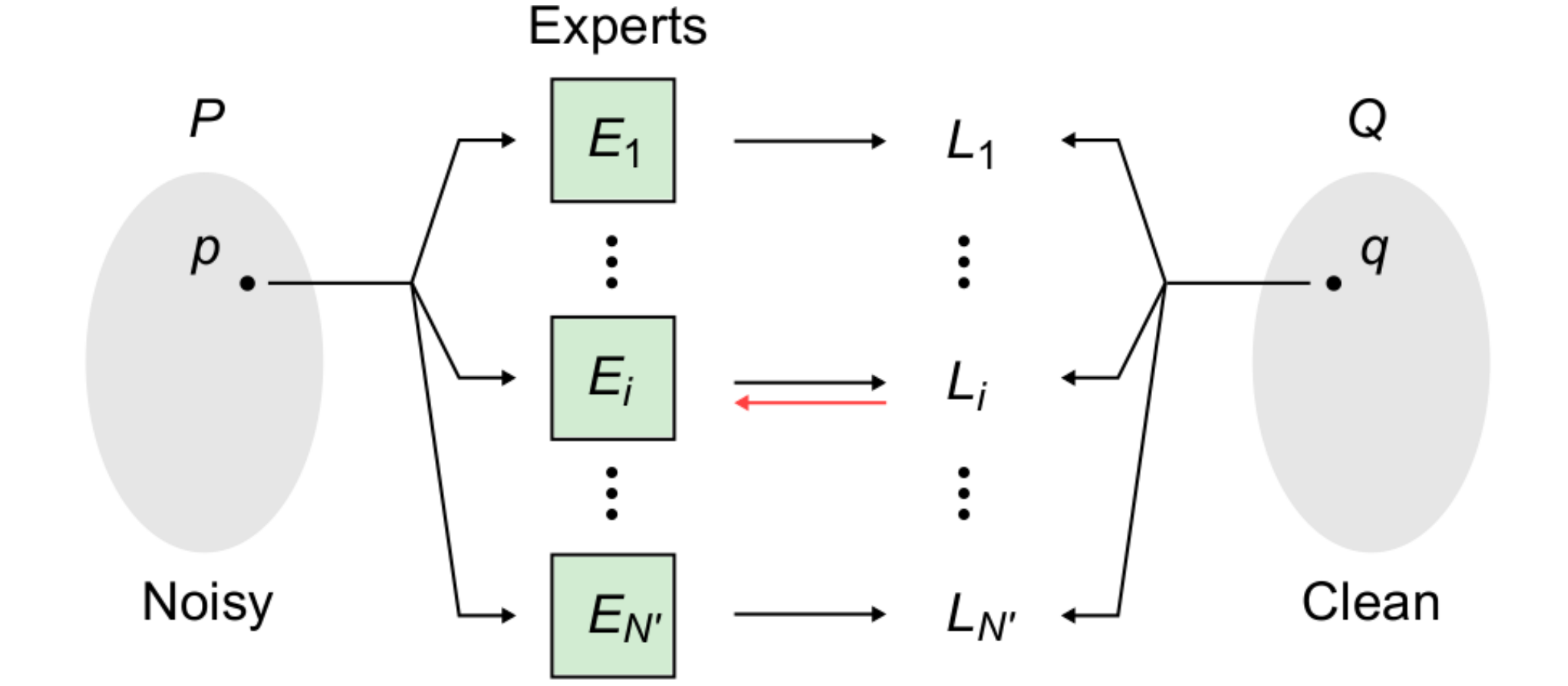}
\caption{Schematic illustration of training procedure with competition of experts.
After preparing experts well trained for dataset $P$ and $Q$, noisy image $p$ is fed to all experts independently, and only parameters of an expert minimizing the loss are updated.}
\label{fig:concept}
\end{figure}

The proposed training procedure is schematically illustrated in Fig.~\ref{fig:concept}.
With given noisy images $P = \{p_i\}^n_{i=1}$ and corresponding ground-truth images $Q = \{q_i\}^n_{i=1}$, our objective is learning a generalized mapping from distribution $\mathcal{D}_P$ to distribution $\mathcal{D}_Q$ without any {\it a priori} information about the noise, where $P \subset \mathcal{D}_P$ and $Q \subset \mathcal{D}_Q$.
Additionally, it is required for practical applications that the mapping is executed with as few parameters as possible.
To achieve this, we train MoE via competition of experts and employ a single network for the evaluation.
The training with the competition consists of a three-step procedure:
\begin{enumerate}
  \item Train all experts $E_1, \cdots, E_{N'}$ for a whole of the dataset $P$ and $Q$ until convergence.
  \item For each iteration, feed noisy images $p$ to all experts independently and update only parameters of an expert minimizing
    \begin{equation}
    \mathcal{L}_j = \mathcal{L}(E_j(p), q),
    \label{eq:loss}
    \end{equation}
  where $\mathcal{L}$ is a given loss function.
  \item Train a gating network with the index of a winning expert $\tilde{j} = \argmin_{1 \le j \le N'}{\mathcal{L}_j}$ as a target label.
  Note that this step can be run in parallel with step~2 because the gating network is independent of each expert.
\end{enumerate}
With this manner, noisy images $P$ are automatically divided into $N$ clusters as a result of the specialization of experts through the catastrophic forgetting~\cite{french1999catastrophic}.
We note that $N' (\ge N)$ is a parameter to be arbitrarily set in advance.
This training procedure is potentially applicable to any network trained with any task dependent loss.

\subsection{Stabilizing Training}
\noindent{\bf{Initialize method.\ }}
An appropriate preparation of experts is important to stabilize the competition.
If we start competition without pre-training (\ie, start from step~2 instead of step~1), the model fails to classify the input dataset.
This is due to that only the expert winning at the initial stage of the competition continues to be updated throughout the subsequent training.
To complete the training properly, we need to prepare experts trained until the convergence for a given dataset (step~1).
We firstly train an expert $E_1$ until convergence, after that just copy its parameters to all other experts $E_2, \cdots, E_{N'}$ to save training cost.
Note that if two or more experts output the same result, we only update the expert with the smallest index among them.\\

\noindent{\bf{Global identity assumption.\ }}
Further stabilizing the competition, we introduce an assumption that any region in a noisy image $p$ belongs to the same domain in the dataset.
We refer to this assumption as {\it global identity assumption}. With this assumption, we randomly sample several patches $\{[p^1, \cdots, p^{N_p}], [q^1, \cdots, q^{N_p}]\}$ from an image pair $\{p, q\}$ and group them into a mini-batch for each iterations, where $N_p$ is a number of patches sampled from an image.
Under the {\it global identity assumption}, Eq.~\ref{eq:loss} is rewritten as
\begin{equation}
\mathcal{L}_j = \sum^{N_p}_{k=1} \mathcal{L}(E_j(p^k), q^k).
\label{eq:loss_gi}
\end{equation}
This assumption is necessary to perform a mini-batch processing in a training with the competition because we assume the input dataset is unlabeled.
We describe an overall procedure of the training with the competition of experts under the {\it global identity assumption} in Algorithm~\ref{alg:training}.

\begin{algorithm}[t]
    \centering
    \caption{Training with the competition of experts}
    \label{alg:training}
    
    {\setlength{\baselineskip}{12.5pt}
    \begin{algorithmic}[1]
    \Require $\{p, q\}$: an image pair sampled from input dataset $P$ and target dataset $Q$, $\{E_i\}^{N'}_{i=1}$: experts with weight parameters $\{\theta_i\}^{N'}_{i=1}$ ($N'$: number of experts), $N_p$: number of patchs sampled from an image, $\mathcal{L}$: loss function
    \Repeat
        \State Sample patches $\{[p^1, \cdots, p^{N_p}], [q^1, \cdots, q^{N_p}]\}$ ($:= \{{\bf p}, {\bf q}\}$) from $\{p, q\}$
        \State Compute $\mathcal{L}_1 = \mathcal{L}(E_1({\bf p}), {\bf q})$
        \State Update $\theta_1$ via backpropagating gradients of $\mathcal{L}_1$
    \Until{Converge}
    \State $\{\theta_i\}^{N'}_{i=2} \Leftarrow \theta_1$
    \Repeat
        \State Sample patches $\{{\bf p}, {\bf q}\}$ from $\{p, q\}$
        \State Compute $\{\mathcal{L}_i\}^{N'}_{i=1} = \{\mathcal{L}(E_i({\bf p}), {\bf q})\}^{N'}_{i=1}$
        \State Let $l = \argmin_{1 \le i \le N'}{\mathcal{L}_i}$\ \ 
        \Comment use a smallest index if duplicates
        \State Update $\theta_l$ via backpropagating gradients of $\mathcal{L}_l$
    \Until{Converge}
    \end{algorithmic}
    }
\end{algorithm}

\begin{figure*}[t]
\centering
\includegraphics[width=0.96\linewidth]{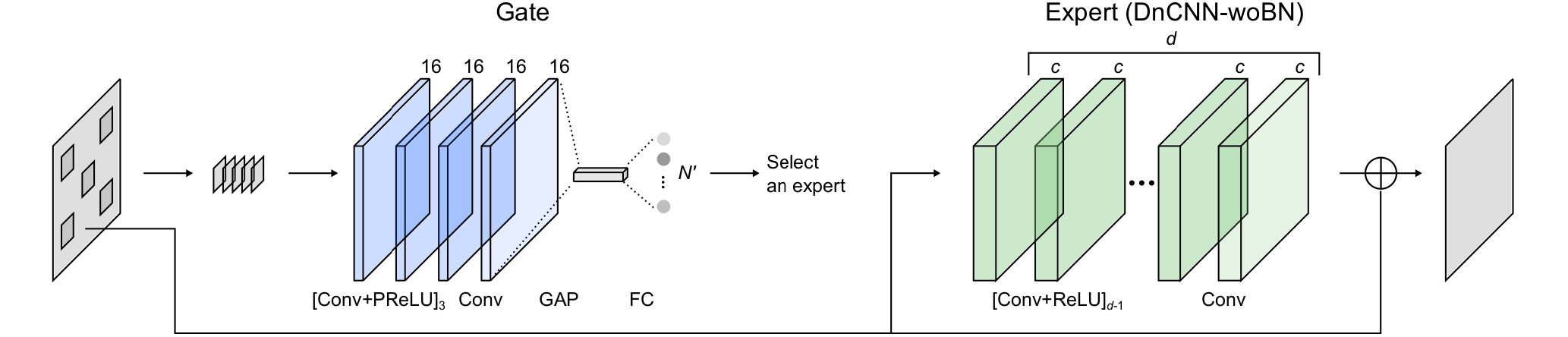}
\caption{Overview of the evaluation network.
Several parts of an input image, which grouped into a mini-batch, are fed into the gating network to select a suitable expert.
After that, the whole of the input image is processed by the selected expert to remove the noise.}
\label{fig:network}
\end{figure*}

\section{Implementation}
\subsection{Network Design}
The expert network used in our experiment is based on DnCNN \cite{zhang2017beyond}.
The each layers of DnCNN are combinations of $3 \times 3$ convolution with zero-padding, ReLU~\cite{glorot2011deep}, and Batch Normalization (BN)~\cite{ioffe2015batch}.
Unlike the original network, we do not utilize BN because we make a mini-batch from a single image in the training dataset (\ie, {\it global identity assumption}), and the mini-batch size is rather small.
In addition, we adopt several different combinations of network depth and number of filters for our experiments.
Here, we refer to our modified version of the original DnCNN as DnCNN-woBN-$d$X$c$Y, where X and Y denote the network depth and the number of filters, respectively.
Note that the original DnCNN consists of 17 and 20 convolution layers with 64 filters for specific noise and blind noise level tasks, respectively.

A gating network which selects a promising expert for an input image is required in the evaluation.
A proposed gating network consists of a plain four-layer CNN followed by a global average pooling (GAP)~\cite{lin2013network} and a fully connected layer.
Each convolution layer has 16 filters with a kernel size of $3 \times 3$, and the first three layers are followed by Parametric ReLU~\cite{he2015delving}.
Since we adopt the {\it global identity assumption} in the training, it is also possible to sample parts of an input image in the evaluation, reducing the computational cost of the gating network.
The gating and expert network architectures are illustrated in Fig.~\ref{fig:network}.

\subsection{Datasets}
We used 400 images from BSDS500~\cite{martin2001database} and 4,744 images of WED~\cite{ma2017waterloo} for the training.
The noisy images were generated by synthesizing AWGNs or JPEG compression noises.
During training, one of the two noise sources was randomly selected with equal probability for an iteration.
The noise level $\sigma$ of AWGN was uniformly selected from a range of [0, 55], and the quality factor $q$ of JPEG compression was uniformly selected from a range of [5, 100].

For the evaluation, 136 images consist of two sets of BSD68~\cite{roth2005fields} was utilized.
One of the test sets was generated by adding AWGNs to the BSD68 with the noise level of $\sigma = \{n \times 55 / 68\}_{n = 0}^{67}$, where $n$ is the index of an image in BSD68 (\ie, the images have different noise levels at an interval of $55/68$).
Similarly for the another test set, quality factors of the JPEG compression were arranged as $q = \{5 + n \times 95 / 68\}_{n = 0}^{67}$.
Here the compression qualities are rounded to integer values.
We refer a test dataset consists of the above two sets of images (\ie, a total of 136 images) to BSD68-GJ.
Note that some of the following evaluations were performed on the plain BSD68 dataset with specific noise levels.

\subsection{Training and Inference}
We cropped 16 patches (the size of each patch is $64 \times 64$) from an image and grouped them into a mini-batch to train the experts ({\it global identity assumption}).
After initializing each expert with an identical pre-trained weight, the experts were specialized via a competition, \ie, only the expert which led to the lowest error for each mini-batch was updated.
During the training, the gating network was simultaneously trained by using the index of the winner expert as the target label.
Owing to this, we can conduct validation at any training stage.
The MSE loss and cross entropy loss were used for training experts and gating network, respectively.
We used Adam optimizer~\cite{kingma2014adam} with a fixed learning rate of $1 \times 10^{-4}$ for both gating and expert networks.
It should be noted that our DnCNN-woBN-$d$17$c$64 trained for a specific level AWGN ($\sigma = 25$) without competition shows similar peak signal-to-noise ratio (PSNR) with the original work: PSNR (ours) = 29.18 dB , PSNR (original work~\cite{zhang2017beyond}) = 29.23 dB.
For a fair comparison, we also trained our model only with BSD400 dataset as in the original work~\cite{zhang2017beyond}. The performance drop was negligible ($-0.01$ dB).

For the evaluation, five non-overlapping patches (the size of each patch was $64 \times 64$) cropped from a test image were fed into the gating network.
Then a suitable expert was selected according to an average of the outputs of the gate.
After that, the selected expert processes the original image to recover a corresponding ground-truth image.
The overall evaluation network is illustrated in Fig.~\ref{fig:network}.

\section{Experiments}
\begin{figure}[t]
\centering
\begin{tabular}{c}
    \begin{minipage}{0.95\hsize}
    \centering
    \includegraphics[width=1.0\linewidth]{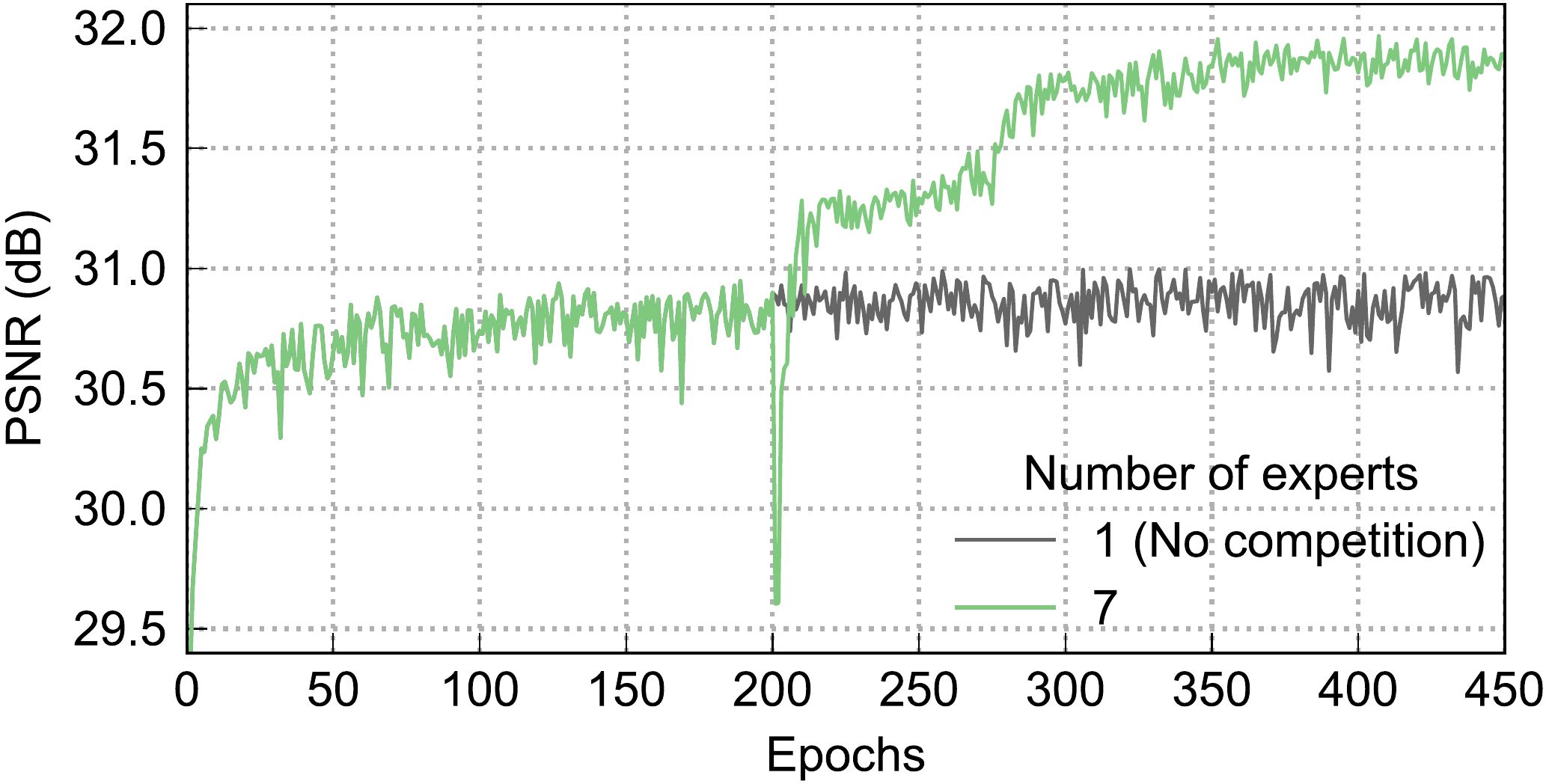}
    \subcaption{}
    \label{fig:psnr}
    \end{minipage}\vspace{3mm}\\

    \begin{minipage}{0.95\hsize}
    \centering
    \includegraphics[width=1.0\linewidth]{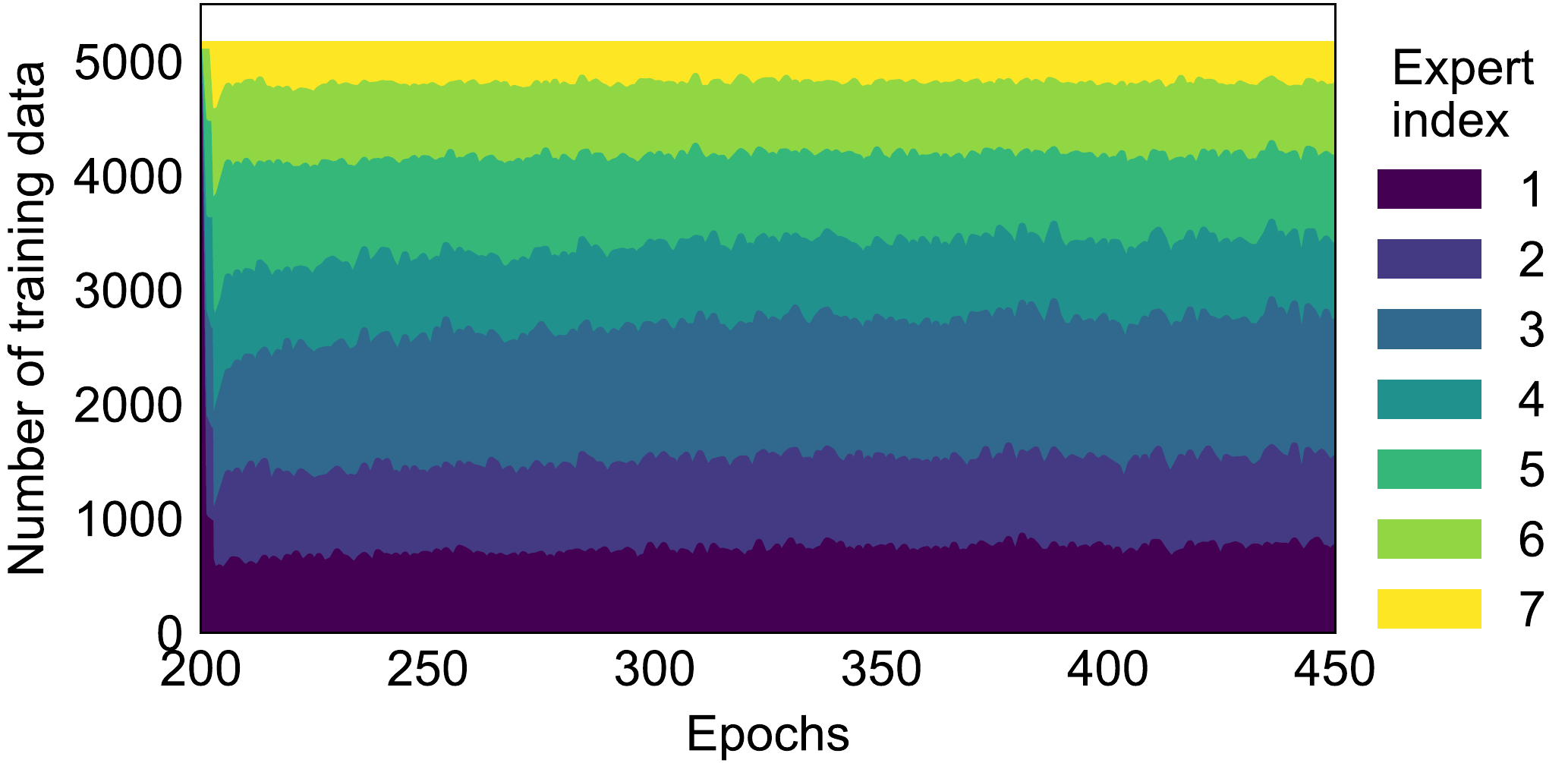}
    \subcaption{}
    \label{fig:domain}
    \end{minipage}
\end{tabular}
\caption{(a) Recorded performance of DnCNN-woBN-$d$5$c$16 for BSD68-GJ obtained with (green) and without (gray) competition.
For the condition with the competition, the number of experts is set to seven and the competition starts from 201st epoch after original expert converges well.
(b) Recorded transitions of winning numbers of the seven experts within the latest epoch.
The experts differentiate into each branch after starting the competition, and the winning numbers finally settle down to a constant ratio.}
\label{fig:train}
\end{figure}

\subsection{Details of Competition of Experts}
Figure~\ref{fig:psnr} shows the historical average PSNR during the training with and without competition.
The DnCNN-woBN-$d$5$c$16 was used as the expert networks, and the PSNR was computed for the BSD68-GJ at each training epoch.
For the condition with the competition, the number of experts $N'$ was set to seven and the competition started from 201st epoch after the original expert converged adequately.
The result demonstrates the effectiveness of the proposed method by about 1 dB improvement in the average PSNR.
Figure~\ref{fig:domain} shows historical transitions of the winning numbers of each expert within the latest epoch.
The experts differentiated into respective branches within 10 epochs after starting the competition.
After that, the winning numbers gradually transited, and finally settle down to a constant ratio.
The PSNR increased in the regime where the ratio of the winning numbers changed, indicating that the training data was spontaneously separated into appropriate domains to enhance the denoising performance.
We note that such a transition behavior during training was reproduced for several executions, showing the stability of the proposed method.

\begin{figure}[t]
\centering
\includegraphics[width=0.96\linewidth]{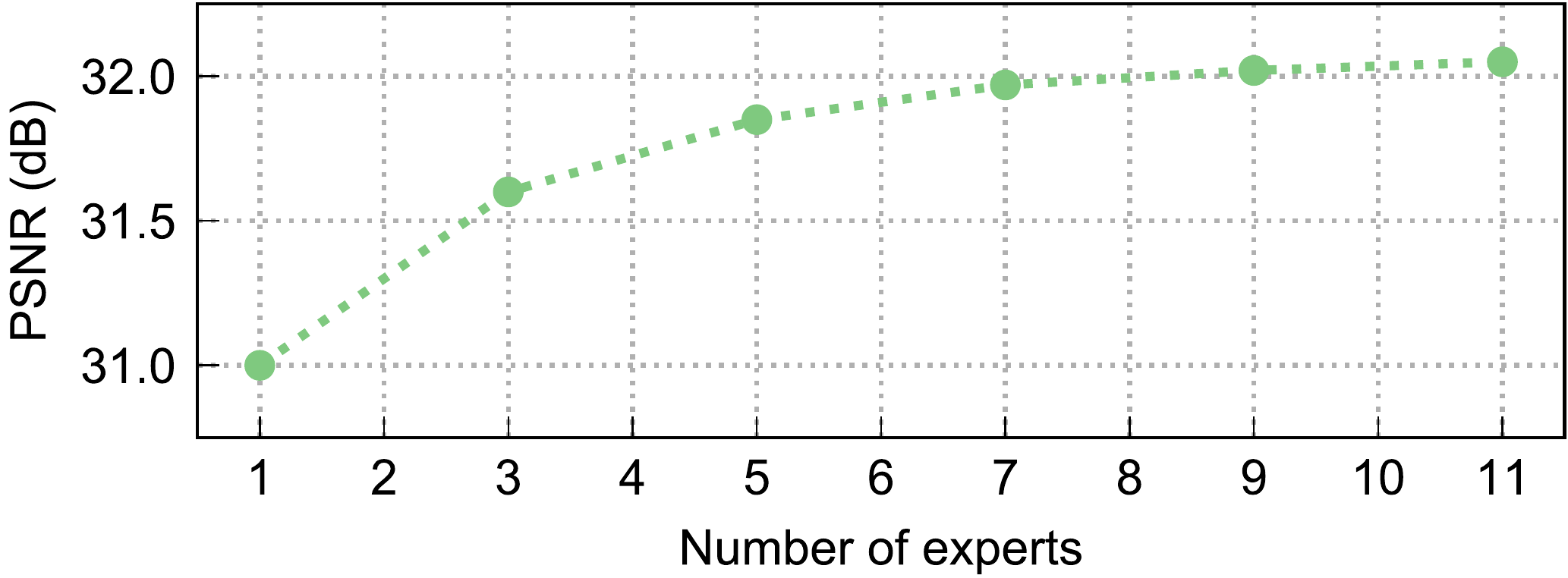}
\caption{Performances of DnCNN-woBN-$d$5$c$16 for BSD68-GJ as a function of the number of experts $N'$.
The average PSNR steadily increases as more experts
are utilized and it almost saturates at seven experts.}
\label{fig:ndepend}
\end{figure}

\begin{figure}[t]
\centering
\includegraphics[width=1.0\linewidth]{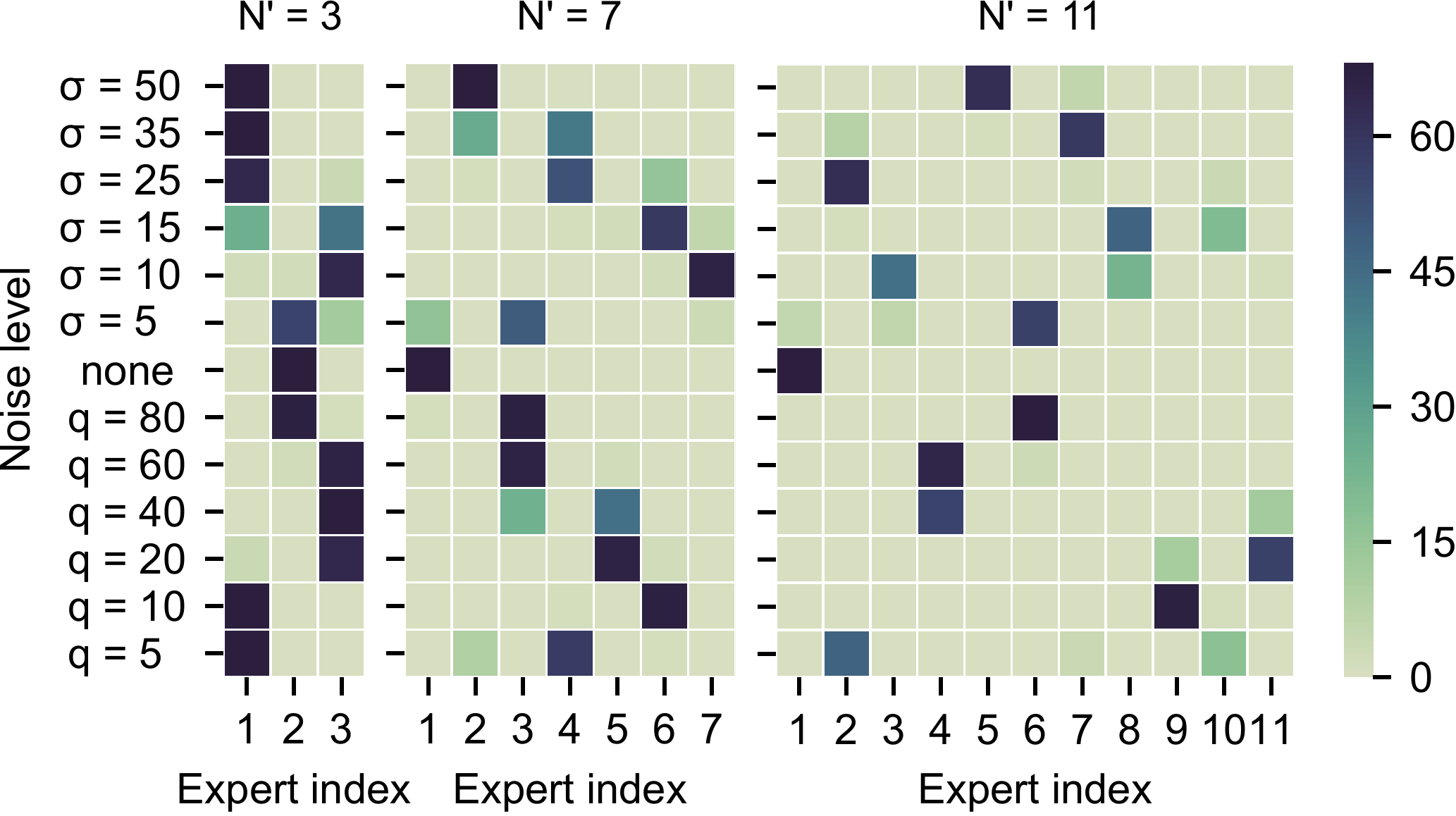}
\caption{Visualization of expert assignment for different noise sources/levels in three conditions of $N' = 3, 7, 11$.
The test images of each row are generated from BSD68 by adding a specific level AWGN ($\sigma = 5, 10, 15, 25, 35, 50$) or compressing via a specific JPEG quality ($q = 80, 60, 40, 20, 10, 5$).
The test images are assigned to an index of expert which maximizes PSNR.}
\label{fig:prop_ndep}
\end{figure}

\subsection{Number of Experts}
Figure~\ref{fig:ndepend} shows the average PSNR of DnCNN-woBN-$d$5$c$16 for the BSD68-GJ as a function of the number of experts $N'$.
The average PSNR steadily increases as more experts are utilized, and it started to seem saturated as the number of experts exceeds seven.
Figure~\ref{fig:prop_ndep} visualizes which expert maximized performance for different noise sources/levels in three conditions of $N' = 3, 7, 11$.
The test images of each row were generated from BSD68 by adding a specific level AWGN ($\sigma = 5, 10, 15, 25, 35, 50$) or compressing via a specific JPEG quality ($q = 80, 60, 40, 20, 10, 5$).
The test images were assigned to an index of expert which maximized PSNR.
In our experiment, training data did not have distinct domains because the noise level was gradated via the random uniform selections.
In consequence, as shown in Fig.~\ref{fig:prop_ndep}, the more experts were implemented, the more finely the training domains were divided.
Such a fine segmentation, however, might trigger overfitting and/or lead to failure in the selection of experts, and thus it might limit denoising performance (see Fig.~\ref{fig:ndepend}).

\begin{figure}[t]
\centering
\includegraphics[width=1.0\linewidth]{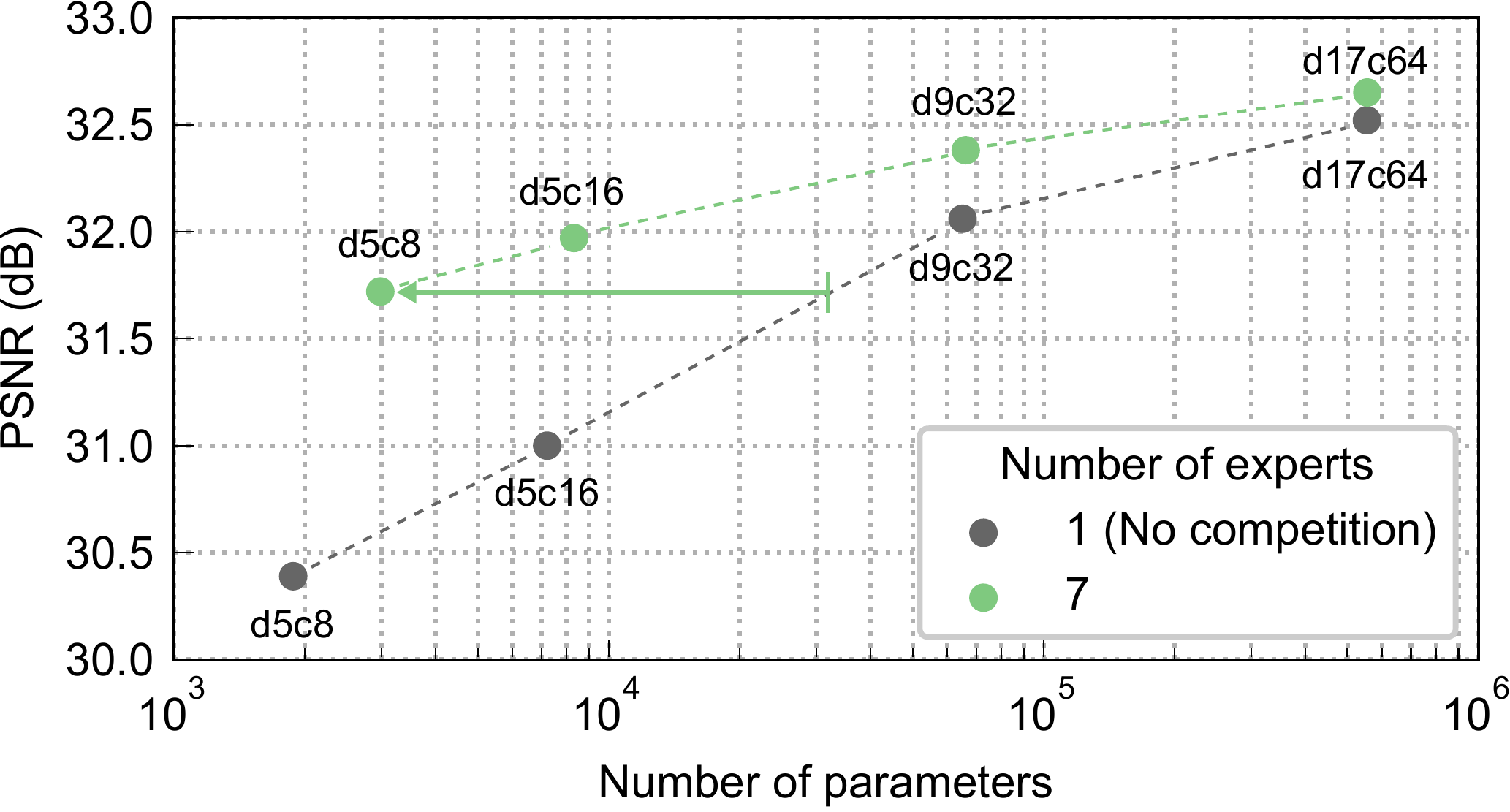}
\caption{PSNR vs. network complexity curve.
Four different expert networks DnCNN-woBN-$d$5$c$8, $d$5$c$16, $d$9$c$32, and $d$17$c$64 are utilized and evaluated for BSD68-GJ with (green) and without (gray) competition.
The number of experts is seven for the condition with the competition.
As depicted by the arrow, the total complexity is reduced by up to approximately $90\%$ without sacrifice of the denoising performance.}
\label{fig:efficiency}
\end{figure}

\subsection{Computational Efficiency}
Figure~\ref{fig:efficiency} shows PSNR vs. network complexity curve (complexity means the total number of parameters).
The four different expert networks DnCNN-woBN-$d$5$c$8, $d$5$c$16, $d$9$c$32, and $d$17$c$64 were utilized in this case and evaluated for BSD68-GJ with and without competition.
The number of experts was seven for the condition with the competition.
The total complexity of this model is defined as a sum of the complexity of the gating network and the expert.
Here the gate complexity is calculated as the product of the original gate complexity and the area ratio of sampled patches to the whole of the input image.
As depicted by the arrow in Fig.~\ref{fig:efficiency}, a total complexity was saved up to approximately $90\%$ by the competition of experts while the accuracy was maintained.
The competition of experts is more efficient and effective than models without competition in the low-complexity regime.
This result indicates our method is suitable for fast processing applications.
Nevertheless, the proposed approach is also beneficial for the d17c64 corresponding to the standard DnCNN (by $+0.13$ dB of PSNR).

\begin{figure*}[t]
\centering
\begin{tabular}{c}
    \begin{minipage}{0.16\hsize}
    \centering
    \includegraphics[width=1.0\linewidth]{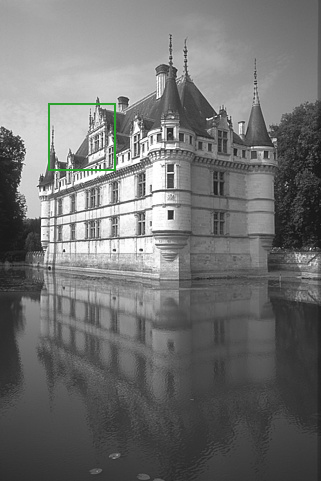}
    \subcaption{Ground truth}
    \end{minipage}

    \begin{minipage}{0.28\hsize}
    \centering
    \includegraphics[width=1.0\linewidth]{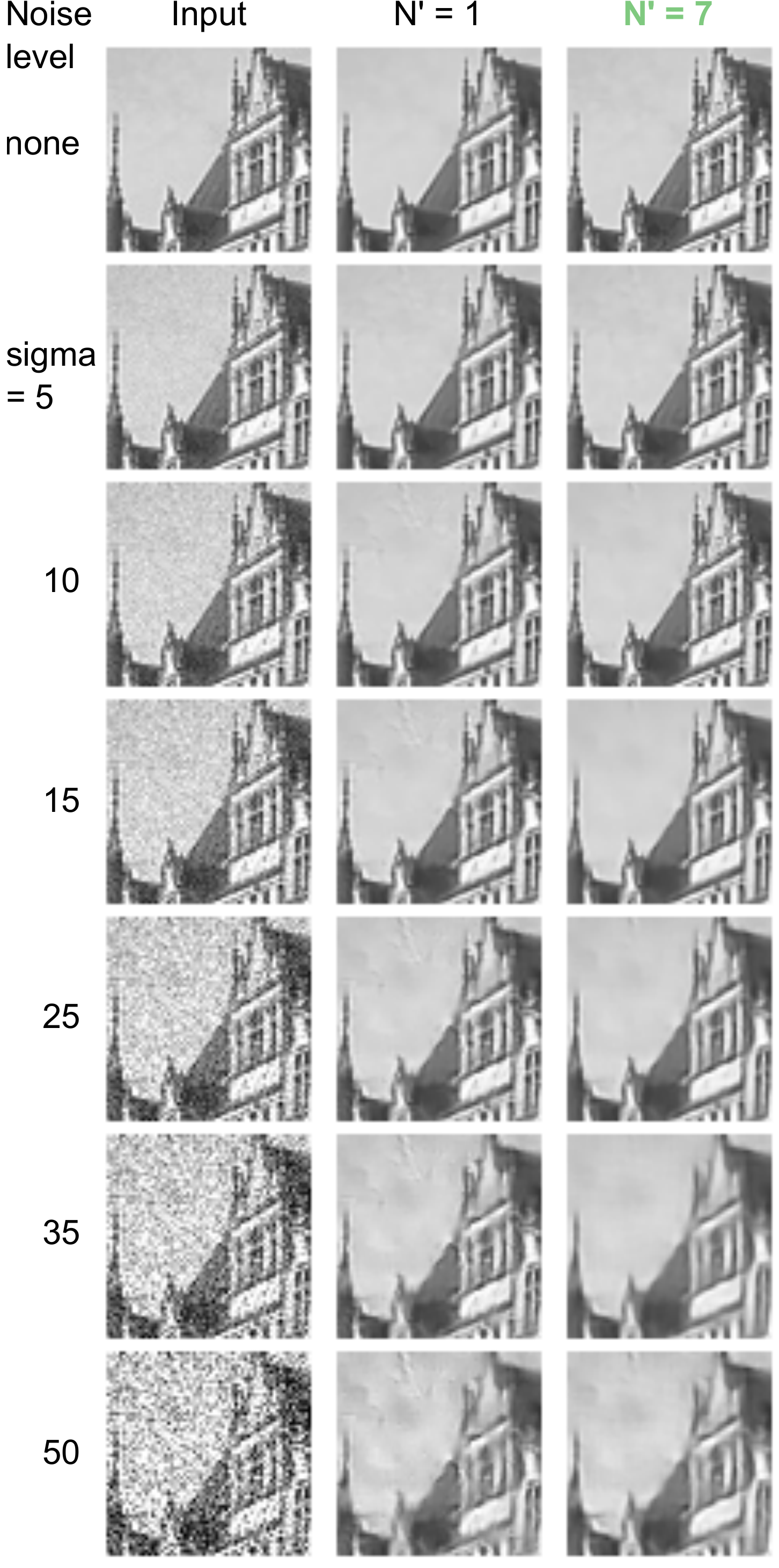}
    \subcaption{AWGN}
    \end{minipage}
    
    \begin{minipage}{0.28\hsize}
    \centering
    \includegraphics[width=1.0\linewidth]{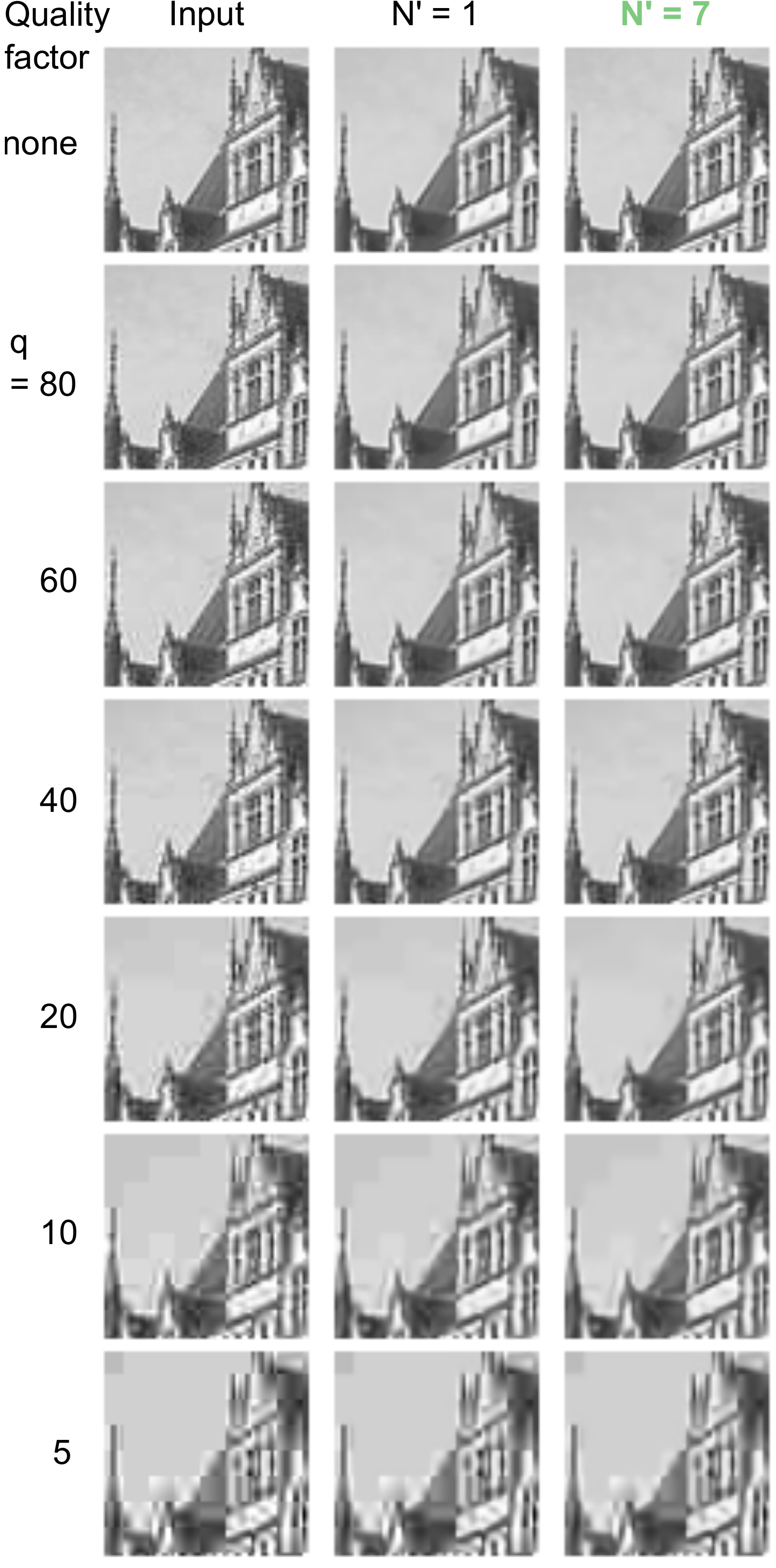}
    \subcaption{JPEG compression}
    \end{minipage}
\end{tabular}
\caption{Visual comparisons between with and without competition.
The employed architecture is DnCNN-woBN-$d$5$c$16, and the number of experts is seven for the competition.
(a) A ground-truth image from BSD68.
(b) From top to bottom: inputs and denoising results with and without competition for different AWGN levels 0 (without noise), 5, 10, 15, 25, 35, and 50, respectively.
(b) From top to bottom: inputs and denoising results with and without competition for different JPEG compression qualities null (without compression), 80, 60, 40, 20, 10, and 5, respectively.}
\label{fig:visual}
\end{figure*}

\subsection{Visual Comparison}
Figure~\ref{fig:visual} shows visual comparisons between with and without competition along with different levels of AWGNs and JPEG compression noises.
The employed architecture was DnCNN-woBN-$d$5$c$16, and the number of experts was seven for the competition.
For almost all noise levels, the results obtained through competition demonstrate better perceptual quality than that of the single model.
In particular, the model trained via competition preserves fine details in the ground-truth images for the low-noise conditions while the single model overly denoises images and tends to blur characteristic details of the ground-truth images.

\begin{table}
\centering
\footnotesize
\begin{tabular}{p{0.95cm}|p{1.31cm} p{1.0cm}|p{1.3cm} p{1.3cm}}
\hline
\hfil Mode & \multicolumn{2}{c|}{Non-Blind} & \multicolumn{2}{c}{Blind (for AWGN and JPEG)}\\
\hline
\hfil Method & \hfil NL-means\footnotemark & \hfil BM3D\footnotemark & \hfil d5c16 & \hfil d9c32\\
\hline
\hfil $\sigma = 15$ & \hfil 29.78 & \hfil 31.10 & \hfil 30.60 & \hfil {\bf 31.15}\\
\hfil $\sigma = 25$ & \hfil 27.43 & \hfil 28.63 & \hfil 28.25 & \hfil {\bf 28.64}\\
\hfil $\sigma = 50$ & \hfil 24.62 & \hfil 25.69 & \hfil 25.25 & \hfil {\bf 25.79}\\
\hline
\hfil Time (s) & \hfil 0.595 & \hfil 4.853 & \hfil {\bf 0.137} & \hfil 0.952\\
\hline
\end{tabular}
\caption{Average PSNR and execution time for BSD68 (image size is $481 \times 321$) synthesized with specific level AWGNs.
The number of experts is seven for our methods DnCNN-woBN-$d$5$c$16 and $d$9$c$32.
All of the algorithms are executed by a single-threaded CPU, and the test images are processed one by one.}
\label{tab:compare_awgn}
\end{table}

\begin{table}
\centering
\footnotesize
\begin{tabular}{p{0.95cm}|p{1.7cm}|p{1.7cm} p{1.7cm} }
\hline
\hfil Mode & \multicolumn{1}{c|}{Blind (for JPEG)} & \multicolumn{2}{c}{Blind (for AWGN and JPEG)}\\
\hline
\hfil Method & \hfil Knusperli & \hfil d5c8 & \hfil d5c16\\
\hline
\hfil $q = 10$ & \hfil 27.14 / 0.7824 & \hfil 27.46 / 0.7874 & \hfil {\bf 27.82} / {\bf 0.7948}\\
\hfil $q = 20$ & \hfil 29.08 / 0.8536 & \hfil 29.86 / 0.8675 & \hfil {\bf 30.18} / {\bf 0.8719}\\
\hfil $q = 40$ & \hfil 31.22 / 0.9041 & \hfil 32.31 / {\bf 0.9176} & \hfil {\bf 32.38} / 0.9162\\
\hline
\hfil Time (s) & \hfil 0.108 & \hfil {\bf 0.068} & \hfil 0.138\\
\hline
\end{tabular}
\caption{Average PSNR, SSIM, and execution time for BSD68 (image size is $481 \times 321$) compressed by specific JPEG qualities.
The number of experts is seven for our methods DnCNN-woBN-$d$5$c$8 and $d$5$c$16.
All of those algorithms are executed by a single-threaded CPU, and the test images are processed one by one.}
\label{tab:compare_jpeg}
\end{table}

\footnotetext[1]{We applied fast-mode NL-means of the Scikit-image with the known noise level $\sigma$ (\ie, non-blind mode).
For each noise levels, the $h$ parameter computing the decay in patch weights was hand-tuned from $0.3 \times \sigma$ to $0.9 \times \sigma$ at interval of $0.1 \times \sigma$ to demonstrate the approximately best-case performance.}
\footnotetext[2]{We used the DCT transform for the step~1 algorithm and the Bior transform for the step~2 algorithm without using any other options. We found this setting demonstrated the best performance in terms of computational efficiency.}

\subsection{Comparison with Prior Arts}
We compare our model ($N' = 7$) trained for both of the AWGN and JPEG noise removal with several existing denoising methods.
As a common benchmark, we used plain BSD68 dataset with specific noise levels.
For AWGN removal, two non-blind patch-based methods were employed as the benchmark: NL-means algorithm implemented in Scikit-image~\cite{scikit-image} and \CC ~implementation of BM3D~\cite{lebrun2012analysis}.
For JPEG compression noise removal, Knusperli~\cite{knusperli} a deblocking JPEG decoder recently developed at Google Open Source was employed as the benchmark.
The execution times of all algorithms were based on a single thread on the laptop machine equipped with an Intel Core i5 CPU @ 2.3GHz with 16GB of memory.

\begin{figure*}[t]
\centering
\begin{tabular}{c}
    \begin{minipage}{0.1268\hsize}
    \centering
    \includegraphics[width=1.0\linewidth]{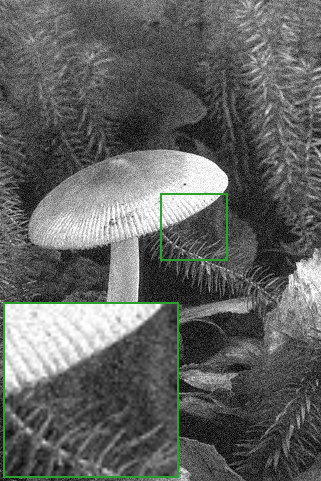}\vspace{1mm}
    \includegraphics[width=1.0\linewidth]{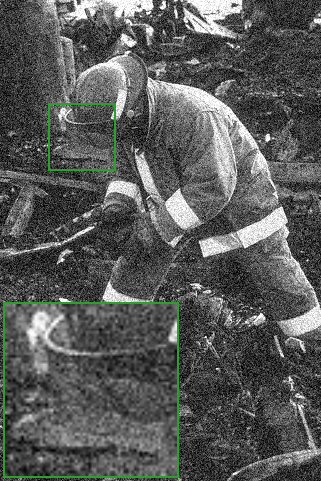}\vspace{1mm}
    \includegraphics[width=1.0\linewidth]{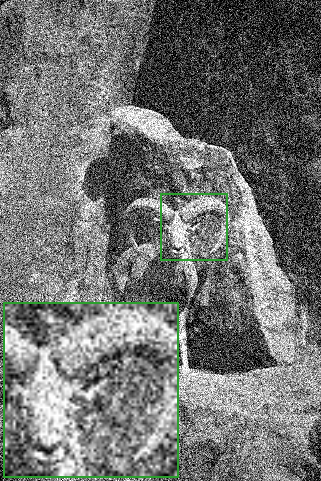}
    \subcaption{Input}
    \end{minipage}

    \begin{minipage}{0.1268\hsize}
    \centering
    \includegraphics[width=1.0\linewidth]{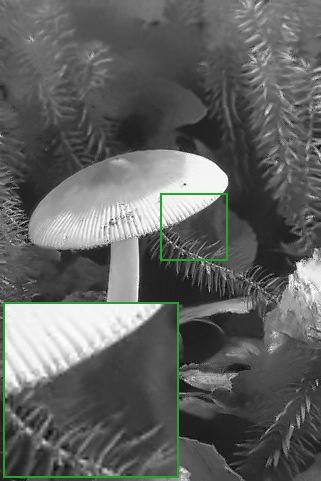}\vspace{1mm}
    \includegraphics[width=1.0\linewidth]{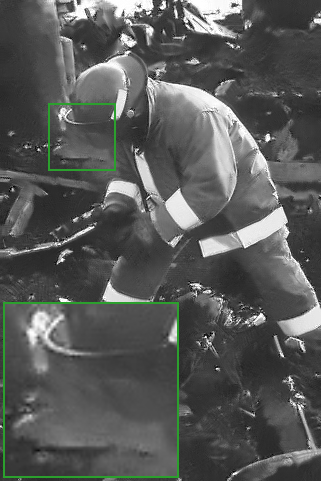}\vspace{1mm}
    \includegraphics[width=1.0\linewidth]{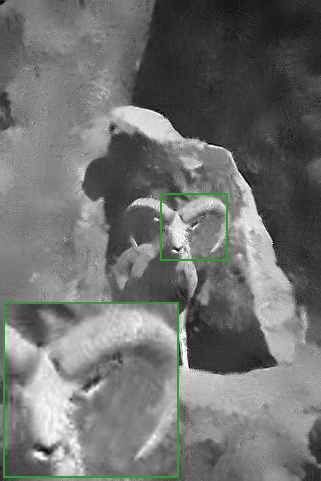}
    \subcaption{NL-means}
    \end{minipage}
    
    \begin{minipage}{0.1268\hsize}
    \centering
    \includegraphics[width=1.0\linewidth]{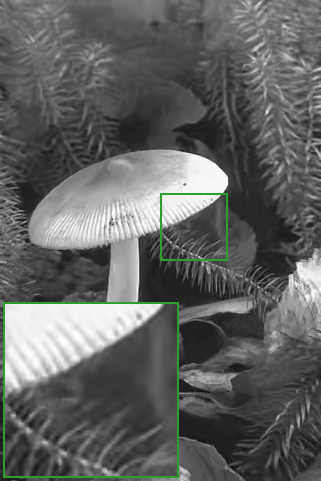}\vspace{1mm}
    \includegraphics[width=1.0\linewidth]{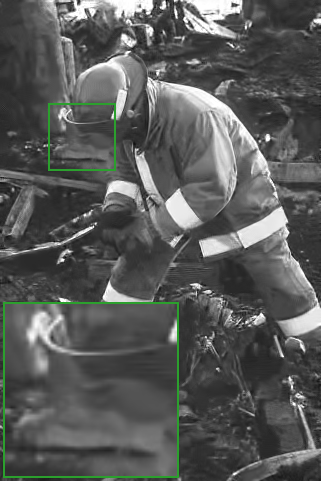}\vspace{1mm}
    \includegraphics[width=1.0\linewidth]{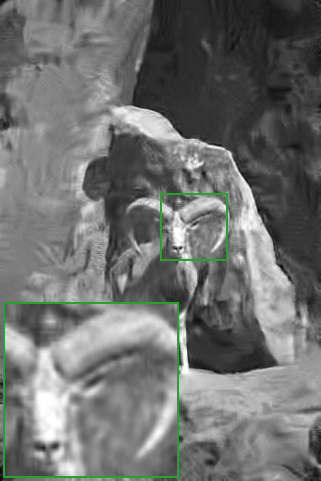}
    \subcaption{BM3D}
    \end{minipage}
    
    \begin{minipage}{0.1268\hsize}
    \centering
    \includegraphics[width=1.0\linewidth]{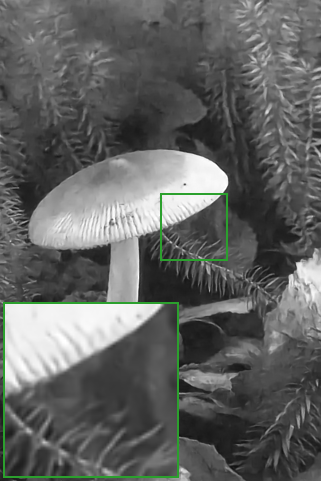}\vspace{1mm}
    \includegraphics[width=1.0\linewidth]{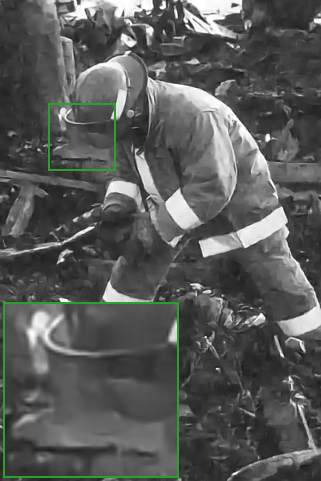}\vspace{1mm}
    \includegraphics[width=1.0\linewidth]{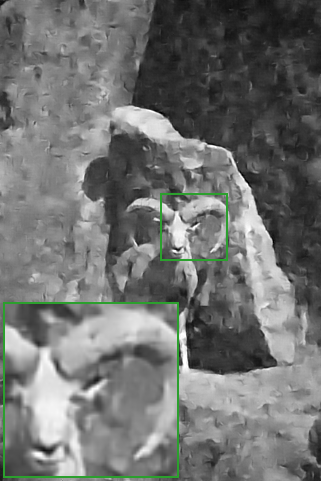}
    \subcaption{d5c16}
    \end{minipage}
    
    \begin{minipage}{0.1268\hsize}
    \centering
    \includegraphics[width=1.0\linewidth]{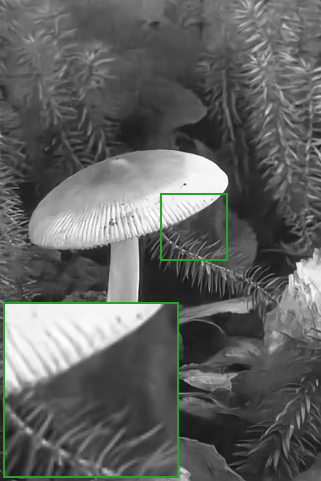}\vspace{1mm}
    \includegraphics[width=1.0\linewidth]{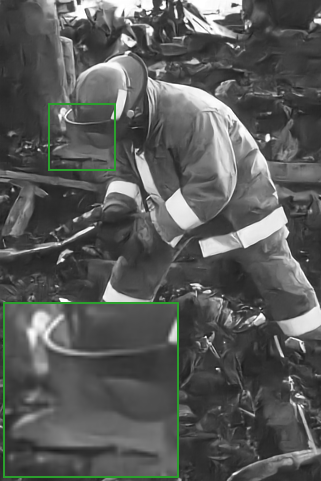}\vspace{1mm}
    \includegraphics[width=1.0\linewidth]{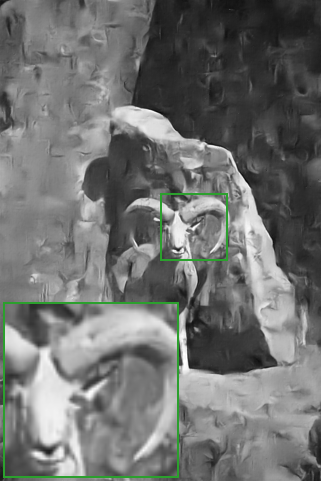}
    \subcaption{d9c32}
    \end{minipage}
    
    \begin{minipage}{0.1268\hsize}
    \centering
    \includegraphics[width=1.0\linewidth]{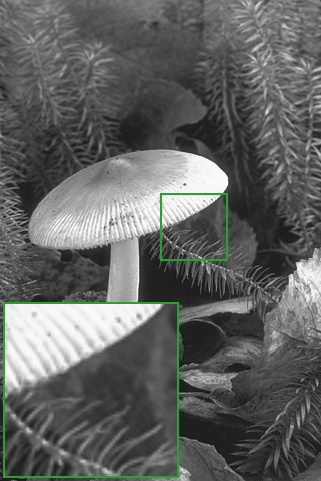}\vspace{1mm}
    \includegraphics[width=1.0\linewidth]{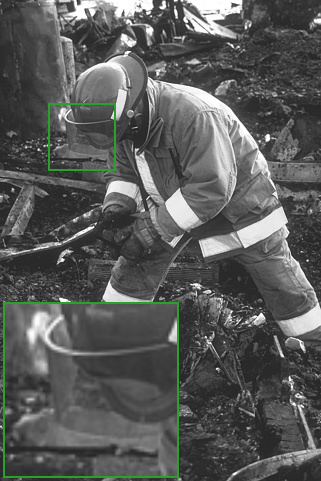}\vspace{1mm}
    \includegraphics[width=1.0\linewidth]{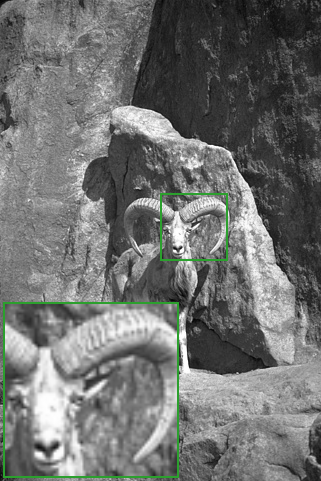}
    \subcaption{Ground truth}
    \end{minipage}
\end{tabular}
\caption{Visual comparisons between our method and prior arts NL-means and BM3D for three test images synthesized with specific level AWGNs.
From top to bottom: $\sigma = 15, 25, 50$.
The number of experts is seven for our methods DnCNN-woBN-$d$5$c$16 and $d$9$c$32.}
\vspace{8mm}
\label{fig:compare_awgn}
\end{figure*}

\begin{figure*}[t]
\centering
\begin{tabular}{c}
    \begin{minipage}{0.19\hsize}
    \centering
    \includegraphics[width=1.0\linewidth]{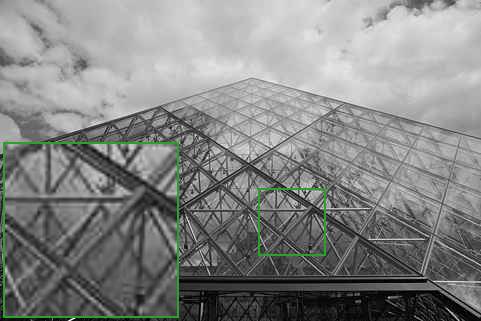}\vspace{1mm}
    \includegraphics[width=1.0\linewidth]{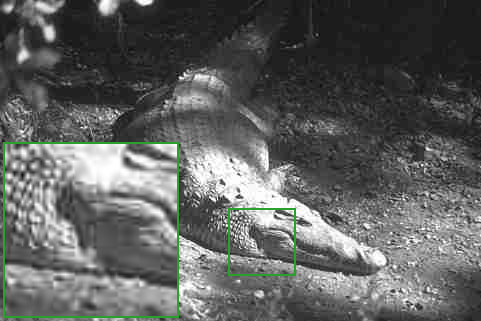}\vspace{1mm}
    \includegraphics[width=1.0\linewidth]{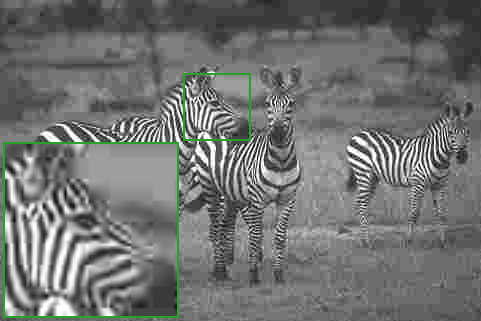}
    \subcaption{Input}
    \end{minipage}

    \begin{minipage}{0.19\hsize}
    \centering
    \includegraphics[width=1.0\linewidth]{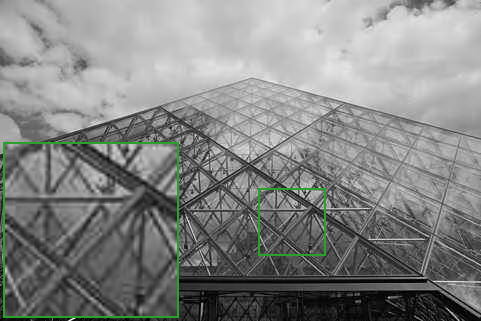}\vspace{1mm}
    \includegraphics[width=1.0\linewidth]{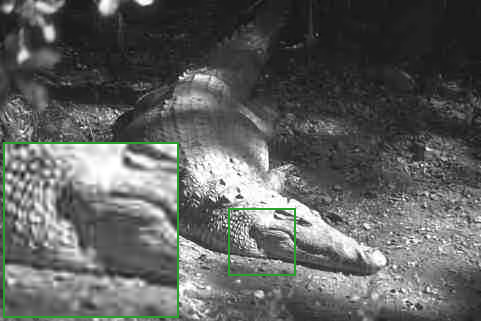}\vspace{1mm}
    \includegraphics[width=1.0\linewidth]{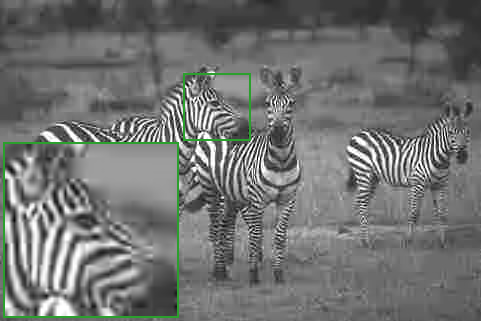}
    \subcaption{Knusperli}
    \end{minipage}
    
    \begin{minipage}{0.19\hsize}
    \centering
    \includegraphics[width=1.0\linewidth]{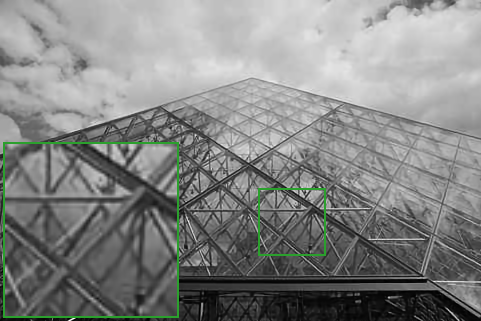}\vspace{1mm}
    \includegraphics[width=1.0\linewidth]{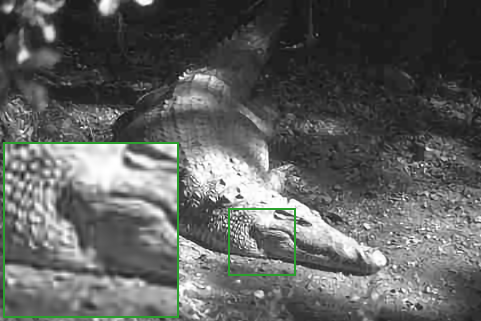}\vspace{1mm}
    \includegraphics[width=1.0\linewidth]{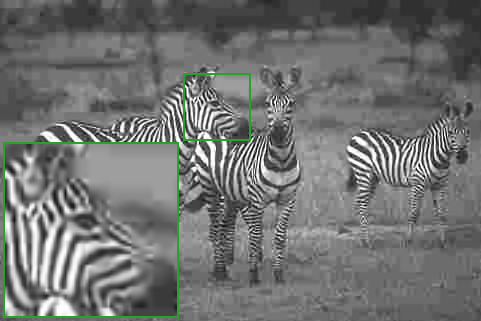}
    \subcaption{d5c8}
    \end{minipage}
    
    \begin{minipage}{0.19\hsize}
    \centering
    \includegraphics[width=1.0\linewidth]{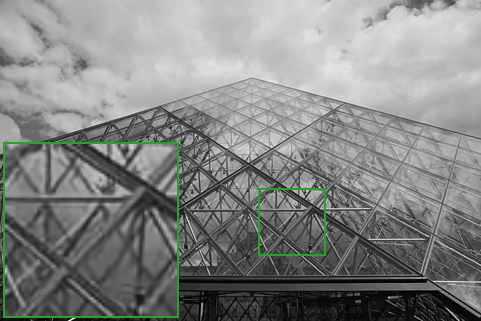}\vspace{1mm}
    \includegraphics[width=1.0\linewidth]{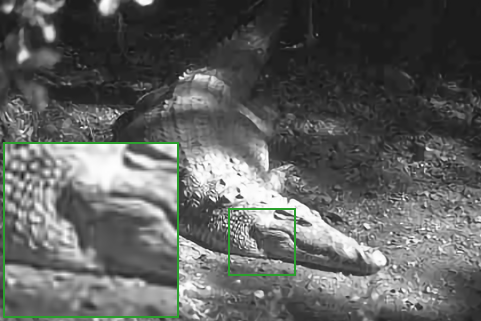}\vspace{1mm}
    \includegraphics[width=1.0\linewidth]{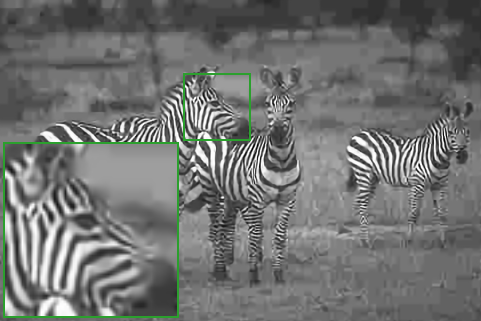}
    \subcaption{d5c16}
    \end{minipage}
    
    \begin{minipage}{0.19\hsize}
    \centering
    \includegraphics[width=1.0\linewidth]{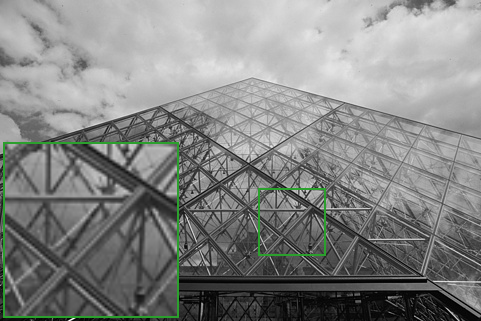}\vspace{1mm}
    \includegraphics[width=1.0\linewidth]{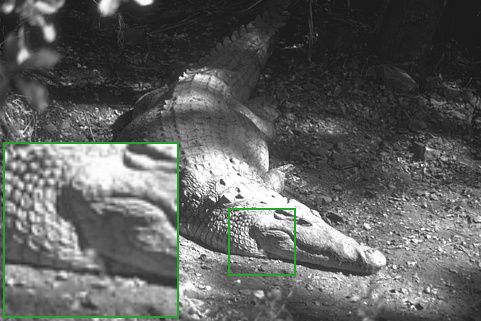}\vspace{1mm}
    \includegraphics[width=1.0\linewidth]{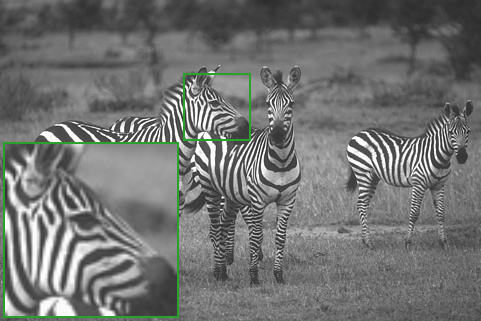}
    \subcaption{Ground truth}
    \end{minipage}
\end{tabular}
\caption{Visual comparisons between our method and a prior art Knusperli for three test images compressed by specific JPEG qualities.
From top to bottom: $q = 40, 20, 10$.
The number of experts is seven for our methods DnCNN-woBN-$d$5$c$8 and $d$5$c$16.}
\label{fig:compare_jpeg}
\end{figure*}

Table~\ref{tab:compare_awgn} summarizes the average PSNR and execution time for BSD68 synthesized with specific level AWGNs.
Our DnCNN-woBN-$d$9$c$32 shows the highest PSNR scores for all of the noise levels, and its execution time is approximately five times faster than that of BM3D.
Furthermore, our DnCNN-woBN-$d$5$c$16 is more than four times faster than fast-mode NL-means while demonstrating overperformance in PSNR by approximately 0.8dB.
Table~\ref{tab:compare_jpeg} summarizes the average PSNR, structural similarity (SSIM)~\cite{wang2004image}, and execution time for BSD68 compressed by specific JPEG qualities.
Both of the DnCNN-woBN-$d$5$c$8 and $d$5$c$16 indicate higher performance than the model of Knusperli, and the execution of DnCNN-woBN-$d$5$c$8 is $37\%$ faster than Knusperli.
Visual comparisons between our method and prior arts for AWGNs and JPEG compression noises are exhibited in Fig.~\ref{fig:compare_awgn} and Fig.~\ref{fig:compare_jpeg}.

\section{Conclusion}
In this paper, we proposed an efficient ensemble network trained via a competition of expert networks, aiming application for low-level vision tasks.
We apply the proposed method to image blind denoising and realize automatic division of unlabeled noisy datasets into clusters respectively optimized to enhance denoising performance.
This method enables to save up to approximately $90\%$ of computational cost without sacrifice of the denoising performance compared to single network models with identical architectures.
We also compare the proposed method with several existing algorithms and observe significant outperformance over prior arts in terms of computational efficiency.

The training procedure of our method relies on an assumption that any region in an input image belongs to the same domain in the dataset.
Consequently, it is difficult to deal with spatially variant low-frequency features of images.
A further extension of our method would be the adaption for other low-level vision tasks.

\section*{Acknowledgement}
I thank Tatsuya Nagata for feedback on drafts of this article. I also thank Hiraku Shibuya and Kazuki Sekine for helpful discussions.

{\small
\bibliographystyle{ieee}
\bibliography{BIB}
}

\end{document}